\documentclass[12pt]{article}

\usepackage[
  margin=1.8cm,
  includefoot,
  footskip=30pt,
]{geometry}

\usepackage{graphicx}
\usepackage{hyperref}
\usepackage{float}
\usepackage{color}
\usepackage{xcolor}
\usepackage{multirow}
\usepackage{array}
\usepackage{threeparttable}
\usepackage{authblk}
\usepackage[symbol]{footmisc}
\usepackage{multicol}
\usepackage{booktabs}
\usepackage{subcaption}
\usepackage{adjustbox}
\usepackage{algorithm}
\usepackage{algpseudocode}
\usepackage{amsmath}
\usepackage{soul}
\usepackage[most]{tcolorbox}

\tcbset{
  aibox/.style={
    width=474.18663pt,
    top=10pt,
    colback=white,
    colframe=black,
    colbacktitle=black,
    enhanced,
    center,
    attach boxed title to top left={yshift=-0.1in,xshift=0.15in},
    boxed title style={boxrule=0pt,colframe=white,},
  }
}
\newtcolorbox{AIbox}[2][]{aibox,title=#2,#1}

\definecolor{Gray}{gray}{0.95}
\definecolor{aigold}{RGB}{244,210, 1} 
\definecolor{aigreen}{RGB}{210,244,211} 

\sethlcolor{aigreen}

\newcommand{\eat}[1]{\ignorespaces}

\title{Can Generalist Foundation Models Outcompete Special-Purpose Tuning?  
Case Study in Medicine}

\author[]{Harsha Nori\textsuperscript{*\textdaggerdbl}}
\author[]{Yin Tat Lee\textsuperscript{*}}
\author[]{Sheng Zhang\textsuperscript{*}}
\author[]{Dean Carignan}
\author[]{Richard Edgar}
\author[]{\newline Nicolo Fusi}
\author[]{Nicholas King}
\author[]{Jonathan Larson}
\author[]{Yuanzhi Li}
\author[]{Weishung Liu}
\author[]{\newline Renqian Luo}
\author[]{Scott Mayer McKinney\textsuperscript{\textdagger}}
\author[]{Robert Osazuwa Ness}
\author[]{\newline Hoifung Poon}
\author[]{Tao Qin}
\author[]{Naoto Usuyama}
\author[]{Chris White}
\author[]{\newline Eric Horvitz\textsuperscript{\textdaggerdbl}}
\affil[]{Microsoft}

\footnotetext[1]{These authors contributed equally.}
\footnotetext[2]{At OpenAI.}
\footnotetext[3]{Correspondence: hanori@microsoft.com, horvitz@microsoft.com}

\date{November 2023}

\begin{document}

\maketitle

\begin{abstract}
Generalist foundation models such as GPT-4 have displayed surprising capabilities in a wide variety of domains and tasks. Yet, there is a prevalent assumption that they cannot match specialist capabilities without intensive training of models with specialty knowledge. For example, most explorations to date on medical competency benchmarks have leveraged domain-specific training, as exemplified by efforts on BioGPT and Med-PaLM. We build on a prior study of the specialist capabilities of GPT-4 on medical challenge benchmarks in the absence of special training. In distinction to the intentional use of simple prompting to highlight the model's out-of-the-box capabilities, we perform a systematic exploration of prompt engineering to boost performance. We find that prompting innovation can unlock deeper specialist capabilities and show that GPT-4 easily tops prior leading results for medical question-answering datasets. 
The prompt engineering methods we explore are general purpose, and make no specific use of domain expertise, removing the need for expert-curated content.  
Our experimental design carefully controls for overfitting during the prompt engineering process. As a culmination of the study, we introduce Medprompt, based on a composition of several prompting strategies. Medprompt greatly enhances GPT-4's performance and achieves state of the art results on all nine of the benchmark datasets in the MultiMedQA suite. The method outperforms state-of-the-art specialist models such as Med-PaLM 2 by a large margin with an order of magnitude fewer calls to the model. 
Steering GPT-4 with Medprompt achieves a 27\% reduction in error rate on the MedQA dataset (USMLE exam) over the best methods to date achieved with specialist models, and surpasses a score of 90\% for the first time. 
Moving beyond medical challenge problems, we show the power of Medprompt to generalize to other domains and provide evidence for the broad applicability of the approach via studies of the strategy on competency exams in electrical engineering, machine learning, philosophy, accounting, law, nursing, and clinical psychology. 
\end{abstract}
\newpage

\section{Introduction}

A long-term aspiration in AI research is to develop principles of computational intelligence and to harness these to build learning and reasoning systems that can perform general problem solving across a diversity of tasks \cite{mccarthy2006proposal,newellshawsimonGPS1959}. In line with this goal, large language models, also referred to as foundation models, such as GPT-3~\cite{brown2020language} and GPT-4~\cite{OpenAI2023GPT4TR}, have demonstrated surprising competencies on a broad swath of tasks without requiring heavy specialized training~\cite{bubeck2023sparks}. These models build on the text-to-text paradigm~\cite{sutskever2014seq2seqmt} with investments in compute and data to learn at scale from indiscriminate consumption of large amounts of public web data. Some of these models are tuned via a learning objective to perform general instruction-following via prompts. 

\begin{figure}[H]
\vspace{10pt}
    \centering
    \begin{subfigure}[t]{0.49\textwidth}
        \centering
        \includegraphics[width=1\textwidth]{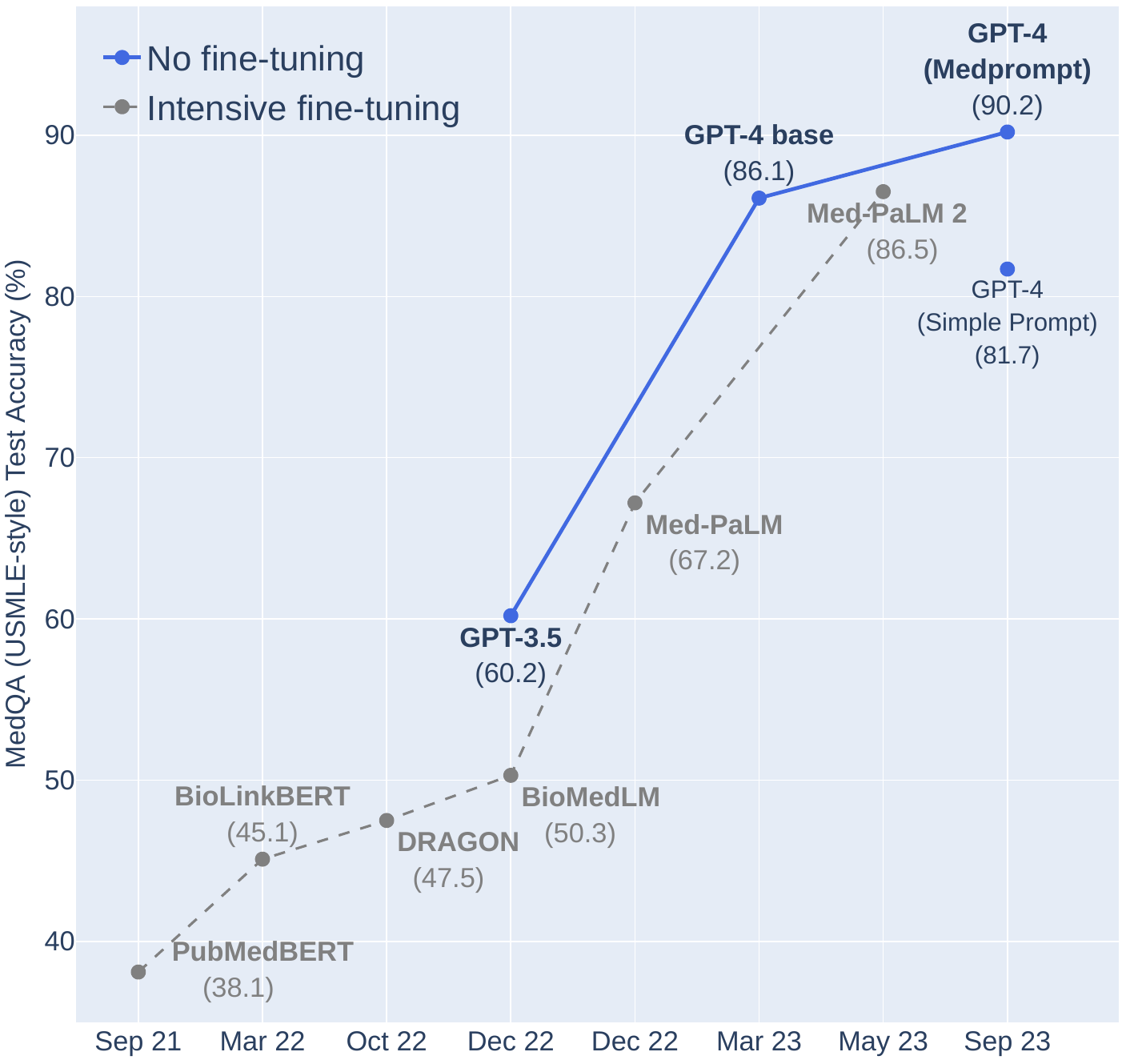}
        \caption{}
        \label{subfig:medqa-comp}
    \end{subfigure}
    \hfill
    \begin{subfigure}[t]{0.49\textwidth}
        \centering
        \includegraphics[width=1\textwidth]{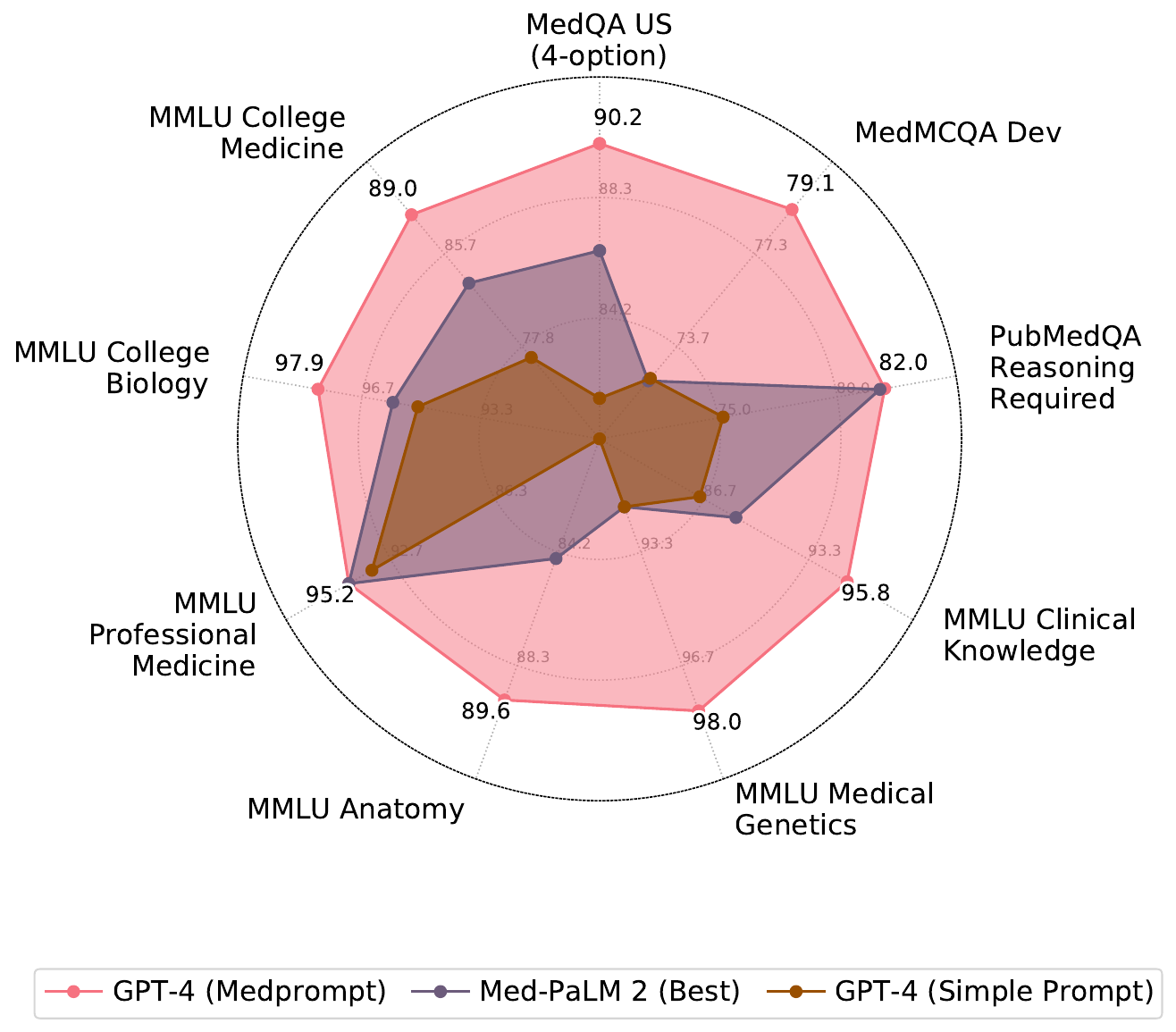}
        \caption{}
        \label{subfig:main-radar}
    \end{subfigure}
    \caption{(a) Comparison of performance on MedQA. (b) GPT-4 with Medprompt achieves SoTA on a wide range of medical challenge questions.}
    \label{fig:comparison}
\end{figure}

A core metric for characterizing the performance of foundation models is the accuracy of next word prediction. Accuracy with next word prediction is found to increase with scale in training data, model parameters, and compute, in accordance with empirically derived ``neural model scaling laws''~\cite{brown2020language,hoffmann2022training}). However, beyond predictions of scaling laws on basic measures such as next word prediction, foundation models show the sudden emergence of numerous problem-solving capabilities at different thresholds of scale~\cite{wei2022emergent,schaeffer2023emergent,OpenAI2023GPT4TR}. 

Despite the observed emergence of sets of general capabilities, questions remain about whether truly exceptional performance can be achieved on challenges within specialty areas like medicine in the absence of extensive specialized training or fine-tuning of the general models. Most explorations of foundation model capability on biomedical applications rely heavily on domain- and task-specific fine-tuning. With first-generation foundation models, the community found an unambiguous advantage with domain-specific pretraining, as exemplified by popular models in biomedicine such as PubMedBERT~\cite{gu2021pubmedbert} and BioGPT~\cite{luo2022biogpt}. But it is unclear whether this is still the case with modern foundation models pretrained at much larger scale.

We focus in this paper on steering foundation models via prompt engineering to excel on a set of medical challenge benchmarks. Med-PaLM 2 attains competitive results on MedQA and other medical challenge problems, via expensive, task-specific fine-tuning of the general PaLM~\cite{chowdhery2022palm} foundation model~\cite{singhal2022large, singhal2023expertlevel}. In addition to reliance on fine-tuning of the base PaLM model, results on the medical benchmarks for Med-PaLM 2 were generated via use of sophisticated, complex prompting strategies, leveraging exemplars crafted by experts. For example, many of the answers rely on an elaborate two-stage prompt scheme of 44 calls for answering each question.

Shortly after GPT-4 was made public in March 2023, several co-authors of this study showed that the model had impressive biomedical competencies ``out-of-the-box'' on medical challenge benchmarks. To demonstrate the latent power of GPT-4 on specialty medical expertise, the co-authors purposefully employed a rudimentary prompting strategy~\cite{nori2023capabilities}. Despite the strong results demonstrated in that study, questions remain about the depth of GPT-4's domain-specific capabilities in the absence of additional special training or tuning. 

We present results and methods of a case study on steering GPT-4 to answer medical challenge questions with innovative prompting strategies. We include a consideration of best practices for studying prompting in an evaluative setting, including the holding out of a true eyes-off evaluation set. We discover that GPT-4 indeed possesses deep specialist capabilities that can be evoked via prompt innovation. The performance was achieved via a systematic exploration of prompting strategies. As a design principle, we chose to explore prompting strategies that were inexpensive to execute and not customized for our benchmarking workload. We converged on a top prompting strategy for GPT-4 for medical challenge problems, which we refer to as Medprompt. Medprompt unleashes medical specialist skills in GPT-4 in the absence of expert crafting, easily topping existing benchmarks for all standard medical question-answering datasets. The approach outperforms GPT-4 with the simple prompting strategy and state-of-the-art specialist models such as Med-PaLM 2 by large margins. On the MedQA dataset (USMLE exam), Medprompt produces a 9 absolute point gain in accuracy, surpassing 90\% for the first time on this benchmark.

As part of our investigation, we undertake a comprehensive ablation study that reveals the relative significance for the contributing components of Medprompt. 
We discover that a combination of methods, including in-context learning and chain-of-thought, can yield synergistic effects. 
Perhaps most interestingly, we find that the best strategy in steering a generalist model like GPT-4 to excel on the medical specialist workload that we study is to use a generalist prompt. We find that GPT-4 benefits significantly from being allowed to design its prompt, specifically with coming up with its own chain-of-thought to be used for in-context learning. This observation echoes other reports that GPT-4 has an emergent self-improving capability via introspection, such as self-verification~\cite{gero2023selfverification}. 

We note that the automated chain-of-thought reasoning removes dependency on special human expertise and medical datasets. Thus, despite the name Medprompt, extending from the framing context and research trajectory of our investigation of the capabilities of GPT-4 on medical challenge problems, the methodology doesn't include any components specifically oriented towards medicine. As we explore in Section \ref{sec:generalization}, the approach can be applied readily to other domains. 
We present details on Medprompt to facilitate future studies on steering generalist foundation models to provide specialist advice.

\section{Background}

\subsection{Foundation Models on Medical Challenge Problems}

In the era of first-generation foundation models, limited model size and computational resources made domain-specific pretraining advantageous. Models such as PubMedBERT~\cite{gu2021pubmedbert}, BioLinkBERT~\cite{yasunaga2022linkbert}, DRAGON~\cite{yasunaga2022dragon}, BioGPT~\cite{luo2022biogpt}, and BioMedLM~\cite{BioMedLM} were pretrained with self-supervised objectives using domain-specific data sources, such as the PubMed corpus and UMLS knowledge graph. Despite their small size and limited computational power, these models demonstrate strong performance in biomedical NLP tasks. More powerful, general-domain foundation models have demonstrated significantly elevated performance in medical challenges without requiring domain-specific pretraining. 

Several studies have explored the performance of generalist foundation models on medical challenge problems. In \cite{kung2023performance}, ChatGPT-3.5 was evaluated on questions drawn from United States Medical Licensing Exam (USMLE), and performed at or near the passing threshold without any specialized training. In \cite{nori2023capabilities}, GPT-4 was shown to exceed the USMLE passing score by over 20 points using simple 5-shot prompting. Other studies have explored the use of foundation models that are specially fine-tuned with medical knowledge. 

Other studies have explored the power of relying on explicit tuning with medical knowledge. Med-PaLM~\cite{singhal2022large} and Med-PaLM 2~\cite{singhal2023expertlevel} leverage fine-tuning of the 540B-parameter Flan-PaLM, using instruction prompt tuning. With Med-PaLM, authors asked a panel of five clinicians to prepare their instruction prompt tuning dataset. Med-PaLM 2, built similarly on PaLM 2, relied on instruction-following full fine-tuning and achieved the state-of-the-art performance on medical QA datasets. 

We re-examine the capabilities of generalist foundation models without resorting to extensive fine-tuning. We explore diverse prompting strategies to best steer powerful generalist foundation models toward delivering strong performance in specialized domains.

\subsection{Prompting Strategies}

\emph{Prompting} in the context of language models refers to the input given to a model to guide the output that it generates. 
Empirical studies have shown that the performance of foundation models on a specific task can be heavily influenced by the prompt, often in surprising ways. For example, recent work shows that model performance on the GSM8K benchmark dataset can vary by over 10\%  without any changes to the model's learned parameters \cite{yang2023large}. \emph{Prompt engineering} refers to the process of developing effective prompting techniques that enable foundation models to better solve specific tasks. Here, we briefly introduce a few key concepts that serve as building blocks for our Medprompt approach.

\emph{In-Context Learning} (ICL) is a key capability of foundation models, allowing the models to solve new tasks from just a few task demonstrations \cite{brown2020language}. For example, an ICL prompt can be created by preceding a test question with several different examples of questions and desired results. ICL does not require updating model parameters but can offer effects similar to fine-tuning. The choice of examples used in few-shot prompting can substantially influence model performance. In our prior investigation of the performance of GPT-4 on medical challenge problems \cite{nori2023capabilities}, we expressly limited prompting to basic in-context learning methods such as fixed one-shot and five-shot prompting to demonstrate the ease with which GPT-4 could be steered to perform with excellence.

\emph{Chain of Thought} (CoT) is a prompting methodology that employs intermediate reasoning steps prior to introducing the sample answer \cite{wei2023chainofthought}. By breaking down complex problems into a series of smaller steps, CoT is thought to help a foundation model to generate a more accurate answer. CoT ICL prompting integrates the intermediate reasoning steps of CoT directly into the few-shot demonstrations. As an example, in the Med-PaLM work, a panel of clinicians was asked to craft CoT prompts tailored for complex medical challenge problems \cite{singhal2022large}. Building on this work, we explore in this paper the possibility of moving beyond reliance on human specialist expertise to mechanisms for generating CoT demonstrations automatically using GPT-4 itself. As we shall describe in more detail, we can do this successfully by providing [question, correct answer] pairs from a training dataset. We find that GPT-4 is capable of autonomously generating high-quality, detailed CoT prompts, even for the most complex medical challenges.

\emph{Ensembling} is a technique for combining the outputs of multiple model runs to arrive at a more robust or accurate result via combining the separate outputs with functions like averaging, consensus, or majority vote. Ensembling methods employing a technique referred to as \emph{self-consistency} \cite{wang2023selfconsistency} use a sampling method to produce multiple outputs that are then consolidated to identify a consensus output. The diversity of the outputs can be controlled by shifting the ``temperature'' parameter in a model's generation, where higher temperatures can be viewed as injecting greater amounts of randomness into the generation process. By reordering or \emph{shuffling} components of a few-shot prompt, ensembling techniques can also address the order sensitivity commonly found with foundation models ~\cite{pezeshkpour2023large, zheng2023large}, thus improving robustness. 

While ensembling can enhance performance, it comes at the cost of increased computational demands. For example, Med-PaLM 2's Ensemble Refinement method used as many as 44 separate inferences for a single question. Due to this computational overhead, we have pursued as a design principle using simpler techniques to avoid excessive inference costs. We report an ablation study in Section \ref{sec:ablation} which explores the potential of further increased performance under increased computational load.

\section{Experimental Design}

We start with an overview of the medical challenge problem datasets and then outline our testing methodology, designed to avoid overfitting that can occur with intensive iteration on a fixed evaluation dataset.

\subsection{Datasets}

Our benchmarks, as reported in Section \ref{sec:results} are primarily based on performance of GPT-4 on 9 multiple-choice, biomedical datasets from the MultiMedQA benchmark suite \cite{singhal2022large}. Specifically, the benchmarks include the following:

\begin{itemize}
    \item \textbf{MedQA} \cite{jin2021disease} contains multiple choice questions in the style of the Medical Licensing Examination questions used to test medical specialist competency in the United States, Mainland China, and Taiwan. For fair comparison with prior work ~\cite{singhal2022large, singhal2023expertlevel, nori2023capabilities}, we focus on the United States subset of the dataset, which has questions in English in the style of the United States Medical Licensing Exam (USMLE). This dataset contains 1273 questions with four multiple choice answers each.
    
    \item \textbf{MedMCQA} \cite{pal2022medmcqa} presents mock and historic exam questions in the style of two Indian medical school entrance exams---the AIIMS and NEET-PG. The ``dev'' subset of the dataset, upon which we report benchmark results (consistent with prior studies), contains 4183 questions, each with four multiple choice answers.
    \item \textbf{PubMedQA} \cite{jin2019pubmedqa} contains tests requiring a yes, no, or maybe answer to biomedical research questions when given context provided from PubMed abstracts. There are two settings for PubMedQA tests called \emph{reasoning-required} and \emph{reasoning-free}. In the reasoning-free setting, a long-form answer that contains explanations of the abstracts is provided. We report results for the reasoning-required setting, in which the model is only given context from abstracts to use when answering the question. This dataset contains a total of 500 questions.

    \item \textbf{MMLU} \cite{hendrycks2020measuring} is a multitask benchmark suite of 57 different datasets spanning domains across STEM, humanities, and social sciences. We follow prior work \cite{singhal2022large} and benchmark against a medically relevant subset of MMLU tasks: clinical knowledge, medical genetics, anatomy, professional medicine, college biology, and college medicine.
\end{itemize}

As we shall see in Section \ref{sec:generalization}, we can test the generality of the Medprompt approach by studying its efficacy for competency exams outside the primary focus on medical challenge problems.  We test our methodology on two nursing datasets focused on answering NCLEX (National Council Licensure Examinaton) questions and six additional datasets from MMLU covering topics like accounting and law. Details of these datasets are presented in Section \ref{sec:generalization}.

\subsection{Sound Testing Methodology} 

While prompting and in-context learning does not change model parameters, a specific choice of prompting strategy can be viewed as a high-level setting or \textit{hyperparameter} of the end-to-end testing process. As a result, we must be cautious about overfitting as part of training and testing, thus providing results that would not generalize out of the training and test sets under consideration. Concerns about overfitting with studies of foundation model performance are similar to the valid concerns in traditional machine learning with overfitting during the hyperparameter optimization process \cite{feurer2019hyperparameter}. We wish to avoid analogous overfitting in the prompt engineering process. 

Intuitively, a prompt harnessing for examples a lookup table of specific benchmark questions will naturally perform much better on those questions than on unseen problems. A common technique to address this problem in traditional machine learning is to create ``test" sets, \emph{which are only evaluated against at the end of the model selection process}.
We adopt this important aspect of sound testing methodology for machine learning studies and randomly carved out 20\% of each benchmark dataset as an ``eyes-off'' split that is completely held out from consideration until the final testing phase. That is, the eyes-off data is kept hidden until the end-stage. The data is not examined or optimized against during the prompt engineering process. For simplicity, we apply the same methodology to every dataset in MultiMedQA, as many of the datasets were not published with dedicated train/test splits by the authors.
In Section \ref{sec:overfitting}, we show the stratified performance of Medprompt on ``eyes-on'' vs. ``eyes-off'' splits of the MultiMedQA datasets. We find that our performance is quite similar between the two, and that GPT-4 with Medprompt actually performs marginally better on the eyes-off, held out data suggesting that the methods will generalize well to similar questions in the ``open world.''  We have not seen evidence of the use of a similar eyes-off approach in prior studies.

\section{Power of Prompting: Exploration and Results}
\label{sec:prompting}

In this section, we detail the three major techniques employed in Medprompt: Dynamic few-shot selection, self-generated chain of thought, and choice shuffle ensembling. After discussing each technique, we review our approach to composing the three methods into the integrated Medprompt.


\subsection{Dynamic Few-shot}
\label{sec:fewshot}

Few-shot learning~\cite{brown2020language} is arguably the most effective in-context learning method.
With the prompting approach, through a few demonstrations, foundation models quickly adapt to a specific domain and learn to follow the task format.
For simplicity and efficiency, the few-shot examples applied in prompting for a particular task are typically fixed; they are unchanged across test examples. This necessitates that the few-shot examples selected are broadly representative and relevant to a wide distribution of text examples. One approach to meeting these requirements is to have domain experts carefully hand-craft \emph{exemplars} \cite{singhal2022large}.
Even so, this approach cannot guarantee that the curated, fixed few-shot examples will be appropriately representative of every test example.
In comparison, when available, the task training set can serve as an inexpensive, high-quality source for few-shot examples.
If the training set is sufficiently large, we can select different few-shot examples for different task inputs. We refer to this approach as employing dynamic few-shot examples. The method makes use of a mechanism to identify examples based 
 on their similarity to the case at hand~\cite{liu2021makes}. For Medprompt, we did the following to identify representative few shot examples:
Given a test example, we choose $k$ training examples that are semantically similar using a $k$-NN clustering in the embedding space.
Specifically, we first use \texttt{text-embedding-ada-002}\footnote{\href{https://openai.com/blog/new-and-improved-embedding-model}{https://openai.com/blog/new-and-improved-embedding-model}} to embed training questions and test questions as vector representations.
Then, for each test question $x$, we retrieve its nearest $k$ neighbors $x_1, x_2, ..., x_k$ from the training set (according to distance in the embedding space of \texttt{text-embedding-ada-002}).
Given a pre-defined similarity measure $d$ such as cosine similarity, the neighbors are ordered in such a way that $d(x_i, x) \leq d(x_j , x)$ when $i < j$.
Compared with fine-tuning, dynamic few-shot leverages the training data, but does not require billions of updates to model parameters.

\subsection{Self-Generated Chain of Thought}
\label{sec:cot}

\begin{figure}[H]
    \centering
    \includegraphics[width=0.75\textwidth]{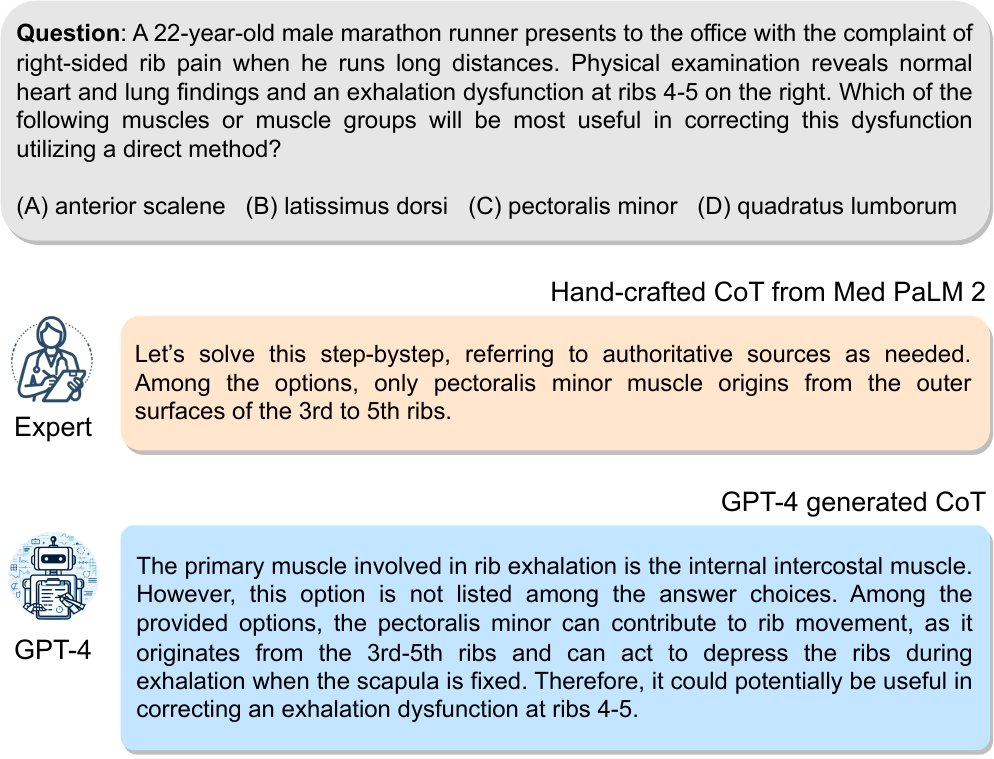}
    \caption{Comparison of expert-crafted and GPT-4-generated chain-of-thought (CoT) prompts. Using a [question, correct answer] pair from a training set, GPT-4 is capable of generating a detailed explanation suitable for use in few-shot CoT demonstrations.}
    \label{fig:cot}
\end{figure}

Chain-of-thought (CoT)~\cite{wei2023chainofthought} uses natural language statements, such as ``\emph{Let's think step by step},'' to explicitly encourage the model to generate a series of intermediate reasoning steps. The approach has been found to significantly improve the ability of foundation models to perform complex reasoning. Most approaches to chain-of-thought center on the use of experts to manually compose few-shot examples with chains of thought for prompting \cite{singhal2023expertlevel}. Rather than rely on human experts, we pursued a mechanism to automate the creation of chain-of-thought examples. We found that we could simply ask GPT-4 to generate chain-of-thought for the training examples using the following prompt:

\begin{figure}[H]
\begin{AIbox}{Self-generated Chain-of-thought Template}
\verb|## Question:|
\verb|{{question}}|
 \\ 
\verb|{{answer_choices}}|
 \\
\verb|## Answer|
 \\
\hl{\textit{model generated chain of thought explanation}} \\
Therefore, the answer is [\hl{\textit{final model answer (e.g. A,B,C,D)}}]
\end{AIbox}
\caption{Template used to prompt foundation model to generate chain-of-thought explanations automatically (detailed in Section \ref{sec:cot}).}
\label{fig:cot-prompt-template}
\end{figure}

\vspace{0.2in} 

A key challenge with this approach is that self-generated CoT rationales have an implicit risk of including hallucinated or incorrect reasoning chains. We mitigate this concern by having GPT-4 generate both a rationale and an estimation of the most likely answer to follow from that reasoning chain.  If this answer does not match the ground truth label, we discard the sample entirely, under the assumption that we cannot trust the reasoning. While hallucinated or incorrect reasoning can still yield the correct final answer (i.e. false positives), we found that this simple label-verification step acts as an effective filter for false negatives. 

We observe that, compared with the CoT examples used in Med-PaLM 2~\cite{singhal2023expertlevel},  which are hand-crafted by clinical experts, CoT rationales generated by GPT-4 are longer and provide finer-grained step-by-step reasoning logic.
Concurrent with our study, recent works~\cite{yang2023large,fernando2023promptbreeder} also find that foundation models write better prompts than experts do.

\subsection{Choice Shuffling Ensemble}
\label{sec:ensemble}
While less severe than other foundation models, GPT-4 can exhibit a propensity to favor certain options in multiple choice answers over others (regardless of the option content), i.e., the model can show position bias~\cite{doi:10.1177/002224378402100210,ko-etal-2020-look,zheng2023judging}.
To reduce this bias, we propose shuffling the choices and then checking consistency of the answers for the different sort orders of the multiple choice.  
As a result, we perform choice shuffle and self-consistency prompting. 
Self-consistency~\cite{wang2023selfconsistency} replaces the naive single-path or \emph{greedy} decoding with a diverse set of reasoning paths when prompted multiple times at some temperature$>0$, a setting that introduces a degree of randomness in generations.
With choice shuffling, we shuffle the relative order of the answer choices before generating each reasoning path.
We then select the most consistent answer, i.e., the one that is least sensitive to choice shuffling.
Choice shuffling has an additional benefit of increasing the diversity of each reasoning path beyond temperature sampling, thereby also improving the quality of the final ensemble \cite{caruana2004ensemble}. We also apply this technique in generating intermediate CoT steps for training examples. For each example, we shuffle the choices some number of times and generate a CoT for each variant. We only keep the examples with the correct answer.

\subsection{Putting it all together: Medprompt}
\label{sec:combined-medprompt}

\begin{figure}[H]
    \centering
    \includegraphics[width=0.95\textwidth]{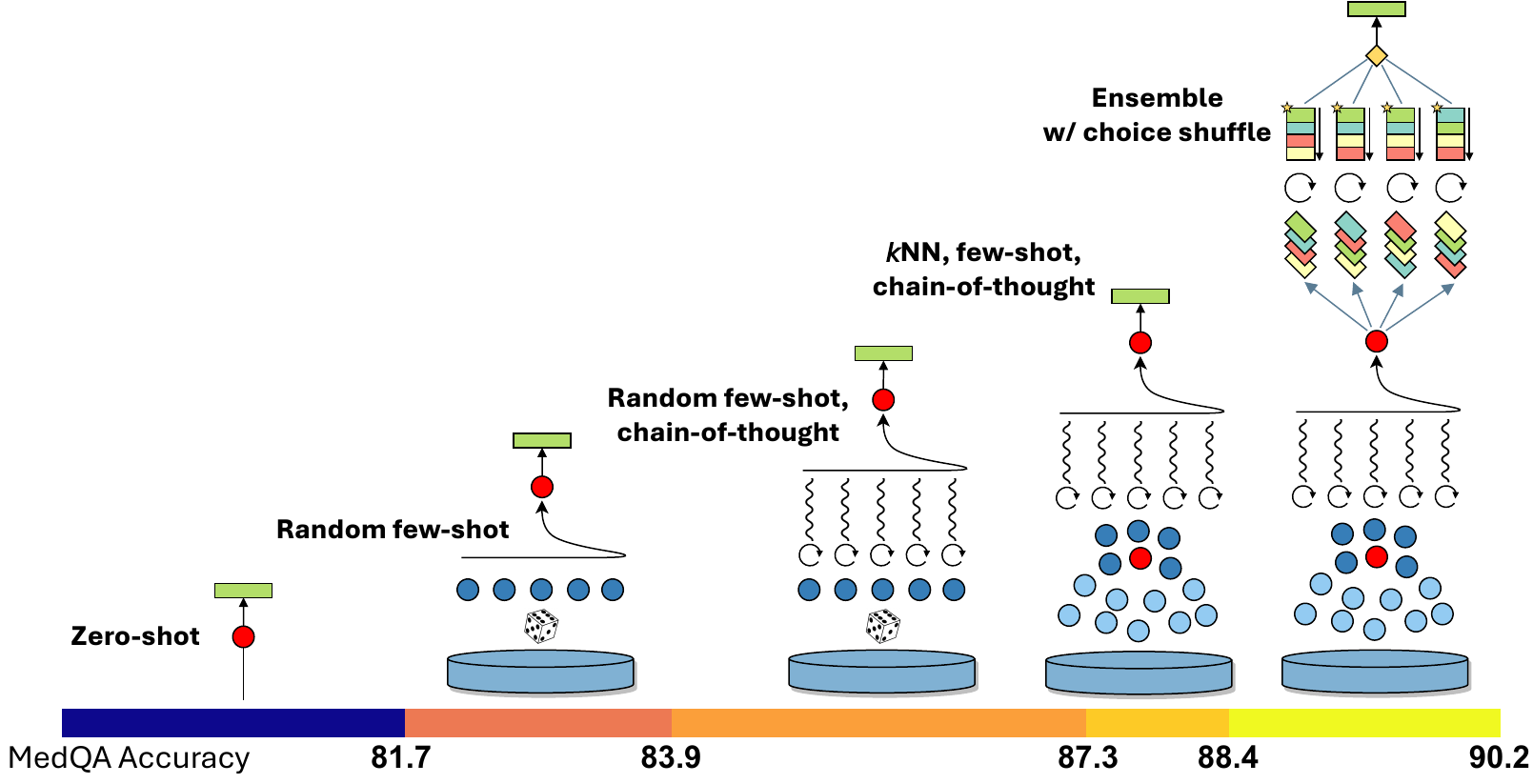}
    \caption{Visual illustration of Medprompt components and additive contributions to performance on the MedQA benchmark. The prompting strategy combines $k$NN-based few-shot example selection, GPT-4--generated chain-of-thought prompting, and answer-choice shuffled ensembling (see details in Section \ref{sec:prompting}). Relative contributions of each component are shown at the bottom (details in Section \ref{sec:ablation}).}
    \label{fig:visual-medprompt}
\end{figure}

Medprompt combines intelligent few-shot exemplar selection, self-generated chain of thought steps, and a majority vote ensemble, as detailed above in Sections \ref{sec:fewshot}, \ref{sec:cot}, and \ref{sec:ensemble}, respectively.  The composition of these methods yields a general purpose prompt-engineering strategy. A visual depiction of the performance of the Medprompt strategy on the MedQA benchmark, with the additive contributions of each component, is displayed in Figure \ref{fig:visual-medprompt}. We provide an a corresponding algorithmic description in Algorithm \ref{alg:medprompt}. 

Medprompt consists of two stages: a \textit{preprocessing} phase and an \textit{inference} step, where a final prediction is produced on a test case. During preprocessing, each question in the training dataset is passed through a lightweight embedding model to generate an embedding vector (Line 4 in Algorithm \ref{alg:medprompt}). We employed OpenAI's text-embedding-ada-002 to create an embedding. For each question, GPT-4 is harnessed to create a chain of thought and a prediction of the final answer (Line 5). If the generated answer is correct and matches the ground truth label, we store the associated question, its embedding vector, the chain of thought, and the answer.  Otherwise, we discard the question entirely from our retrieval pool, with the assumption that we cannot trust the reasoning if the model ultimately arrives at the wrong final answer (Lines 6-7). 

At inference time, given a test question, we re-embed the test sample with the same embedding model used during pre-processing, and utilize $k$NN to retrieve similar examples from the preprocessed pool (Lines 12-13). 
These examples, and their corresponding GPT-4 generated reasoning chains, are structured as context for GPT-4 (Line 14). The test question and corresponding answer choices are then appended at the end, which serves as the final prompt (Line 17). The model, following the few shot exemplars, then outputs a chain of thought and a candidate answer.
Finally, we perform an ensembling process consisting of repeating the steps described above multiple times. We increase diversity by shuffling the answer choices of the test question (Lines 15-16), as detailed in Section \ref{sec:ensemble} and Figure \ref{fig:visual-medprompt}. To determine the final predicted answer, we select the most frequent answer (Line 20).

\begin{algorithm}
\caption{Algorithmic specification of Medprompt, corresponding to the visual representation of the strategy in Figure \ref{fig:visual-medprompt}.}
\label{alg:medprompt}
\begin{algorithmic}[1]
\State \textbf{Input}: Development data $\mathcal{D}$, Test question $Q$
\State \textbf{Preprocessing}:
\For{\texttt{each question} $q$ \texttt{in} $\mathcal{D}$}
    \State Get an embedding vector $v_q$ for $q$.
    \State Generate a chain-of-thought $C_q$ and an answer $A_q$ with the LLM.
    \If{Answer $A_q$ is correct}
        \State Store the embedding vector $v_q$, chain-of-thought $C_q$, and answer $A_q$.
    \EndIf
\EndFor
\State 
\State \textbf{Inference Time}:
\State Compute the embedding $v_Q$ for the test question $Q$.
\State Select the 5 most similar examples $\{(v_{Q_i}, C_{Q_i}, A_{Q_i})\}_{i=1}^5$ from the preprocessed training data using KNN, with the distance function as the cosine similarity: $\text{dist}(v_q, v_Q) = 1 - \frac{\langle v_q, v_Q \rangle}{\|v_q\| \|v_Q\|}$.
\State Format the 5 examples as context $\mathcal{C}$ for the LLM.
\For{\texttt{5 times}}
    \State Shuffle the answer choices of the test question.
    \State Generate a chain-of-thought $C_q^k$ and an answer $A_q^k$ with the LLM and context $\mathcal{C}$.
\EndFor
\State Compute the majority vote of the generated answers $\{A_q^k\}_{k=1}^K$:
\[
    A^{\text{Final}} = \operatorname{mode}(\{A_q^k\}_{k=1}^K),
\]
where $\operatorname{mode}(X)$ denotes the most common element in the set $X$.
\State \textbf{Output}: Final answer $A^{\text{Final}}$.
\end{algorithmic}
\end{algorithm}

The Medprompt results we report here are configured to use \textbf{5} $k$NN selected few shot exemplars and \textbf{5} parallel API calls as part of the choice-shuffle ensemble procedure, which we find strikes a reasonable balance between minimizing inference cost and maximizing accuracy.

Our ablation studies, detailed in Section \ref{sec:ablation}, suggest that further improvements may be achieved by increasing these hyperparameter values. For example, by increasing to 20 few-shot exemplars and 11 ensemble items, we achieve a further $+0.4\%$ performance on MedQA, setting a new state-of-the-art performance threshold of $\mathbf{90.6\%}$.

We note that, while Medprompt achieves record performance on medical benchmark datasets, the algorithm is general purpose and is not restricted to the medical domain or to multiple choice question answering. We believe the general paradigm of combining intelligent few-shot exemplar selection, self-generated chain of thought reasoning steps, and majority vote ensembling can be broadly applied to other problem domains, including less constrained problem solving tasks (see Section \ref{sec:generalization} for details on how this framework can be extended beyond multiple choice questions).

\section{Results}
\label{sec:results}

\begin{table}[H]
\begin{center}
\begin{threeparttable}
\centering
\caption{Performance of different foundation models on multiple choice components of MultiMedQA \cite{singhal2022large}. GPT-4 with Medprompt outperforms all other models on every benchmark.}

    \begin{tabular}{lcccc}
        \toprule 
        \multirow{2}{*}{Dataset}  & \multicolumn{1}{c}{Flan-PaLM 540B\tnote{*}} & \multicolumn{1}{c}{Med-PaLM 2\tnote{*}} & \multicolumn{1}{c}{GPT-4} & \multicolumn{1}{c}{GPT-4} \\
        & \multicolumn{1}{c}{(choose best)} & \multicolumn{1}{c}{(choose best)} & \multicolumn{1}{c}{(5 shot)} & \multicolumn{1}{c}{(Medprompt)}\\ 
        \midrule 
        \textbf{MedQA} & & & & \\
        US (4-option) & 67.6 & 86.5 & 81.4 & \textbf{90.2}\tnote{**}  \\
        \midrule 
        \textbf{PubMedQA} & & & & \\
        Reasoning Required & 79.0 & 81.8 & 75.2 & \textbf{82.0} \\
        \midrule 
        \textbf{MedMCQA} & & & & \\
        Dev & 57.6 & 72.3 & 72.4 & \textbf{79.1} \\
        \midrule 
        \textbf{MMLU} & & & & \\ 
        Clinical Knowledge & 80.4 & 88.7 & 86.4 & \textbf{95.8} \\
        Medical Genetics & 75.0 & 92.0 & 92.0 & \textbf{98.0} \\
        Anatomy & 63.7 & 84.4 & 80.0 & \textbf{89.6} \\
        Professional Medicine & 83.8 & \textbf{95.2} & 93.8 & \textbf{95.2} \\
        College Biology & 88.9 & 95.8 & 95.1 & \textbf{97.9} \\ 
        College Medicine & 76.3 & 83.2 & 76.9 & \textbf{89.0} \\
        \bottomrule 
    \end{tabular}
    \begin{tablenotes}
        \item[*] Sourced directly from \cite{singhal2022large} and \cite{singhal2023expertlevel}. ``Choose best" refers to a process used in the Med-Palm studies of executing several distinct approaches and selecting the best performing strategy for each dataset among the variety of experimental methods tried. Flan-PaLM 540B and Med-PaLM 2 are also both fine-tuned on subsets of these benchmark datasets. By contrast, every GPT-4 reported number uses a single, consistent strategy across all datasets.
        \item[**] We achieve 90.6\%, as discussed in Section \ref{sec:ablation}, with $k=20$ and 11x ensemble steps. The 90.2\% represents ``standard" Medprompt performance with $k=5$ few shot examples and a 5x ensemble.
    \end{tablenotes}

\label{tab:multimedQA-scores}
\end{threeparttable} \\
\end{center}
\end{table}

With harnessing the prompt engineering methods described in Section \ref{sec:prompting} and their effective combination as Medprompt, GPT-4 achieves state-of-the-art performance on every one of the nine benchmark datasets in MultiMedQA.

\subsection{Performance on Eyes-Off Data}
\label{sec:overfitting}

\begin{figure}[H]
    \centering
    \includegraphics[width=\textwidth]{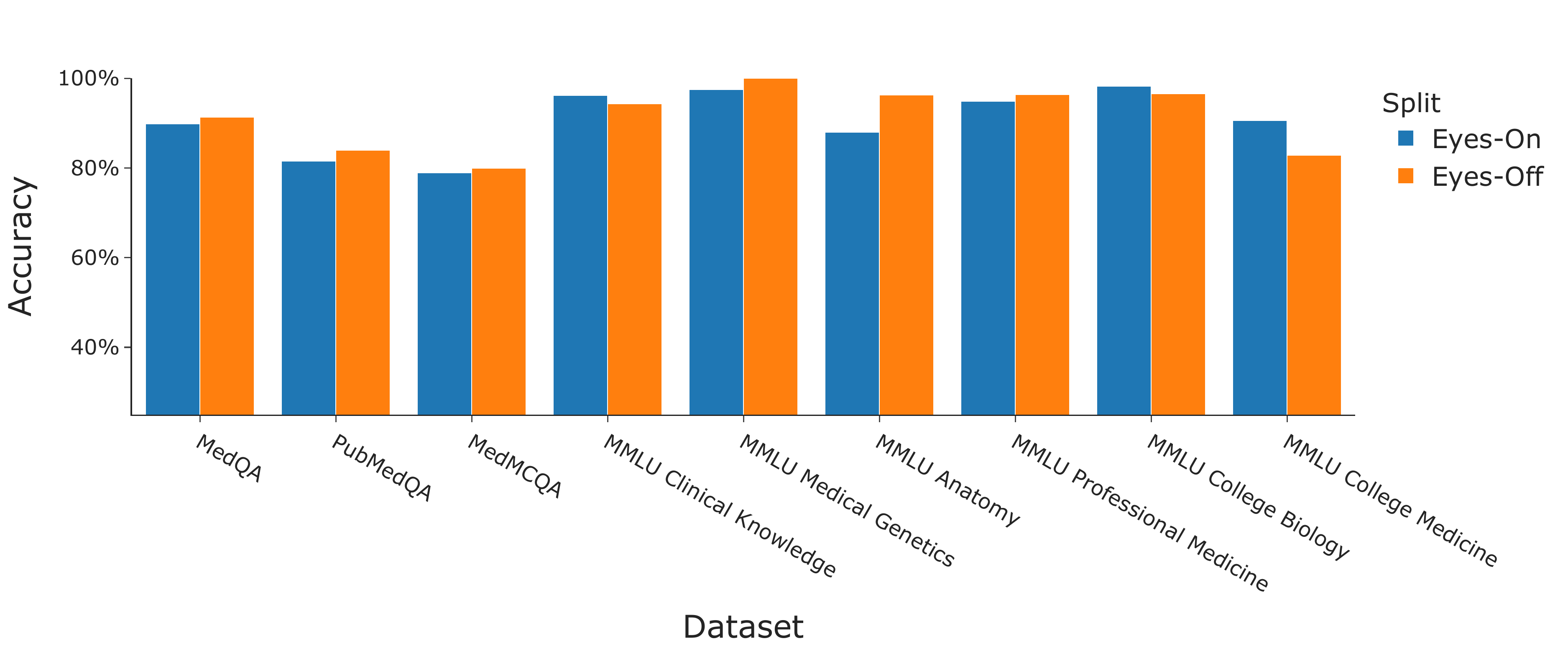}
    \caption{Medprompt evaluation against 20\% eyes-off holdout. Medprompt performs better on the eyes-off dataset in the majority of cases.
    }
    \label{fig:overfitting}
\end{figure}

As introduced in Section \ref{sec:overfitting}, we evaluated the Medprompt prompting design on a held-out ``eyes-off'' subset of each benchmark dataset to check for overfitting risk. GPT-4 with Medprompt achieved an average performance of 90.6\% on the eyes-on data, and an average performance of 91.3\% on the eyes-off data, suggesting that the prompt engineering process likely did not lead to overfitting on MultiMedQA datasets. As additional evidence, the performance on eyes-off data was better in 6/9 of the benchmark datasets (Figure \ref{fig:overfitting}).

\subsection{Insights about Medprompt Components via Ablation Studies}
\label{sec:ablation}

\begin{figure}[H]
    \centering
    \includegraphics[width=1.0\textwidth]{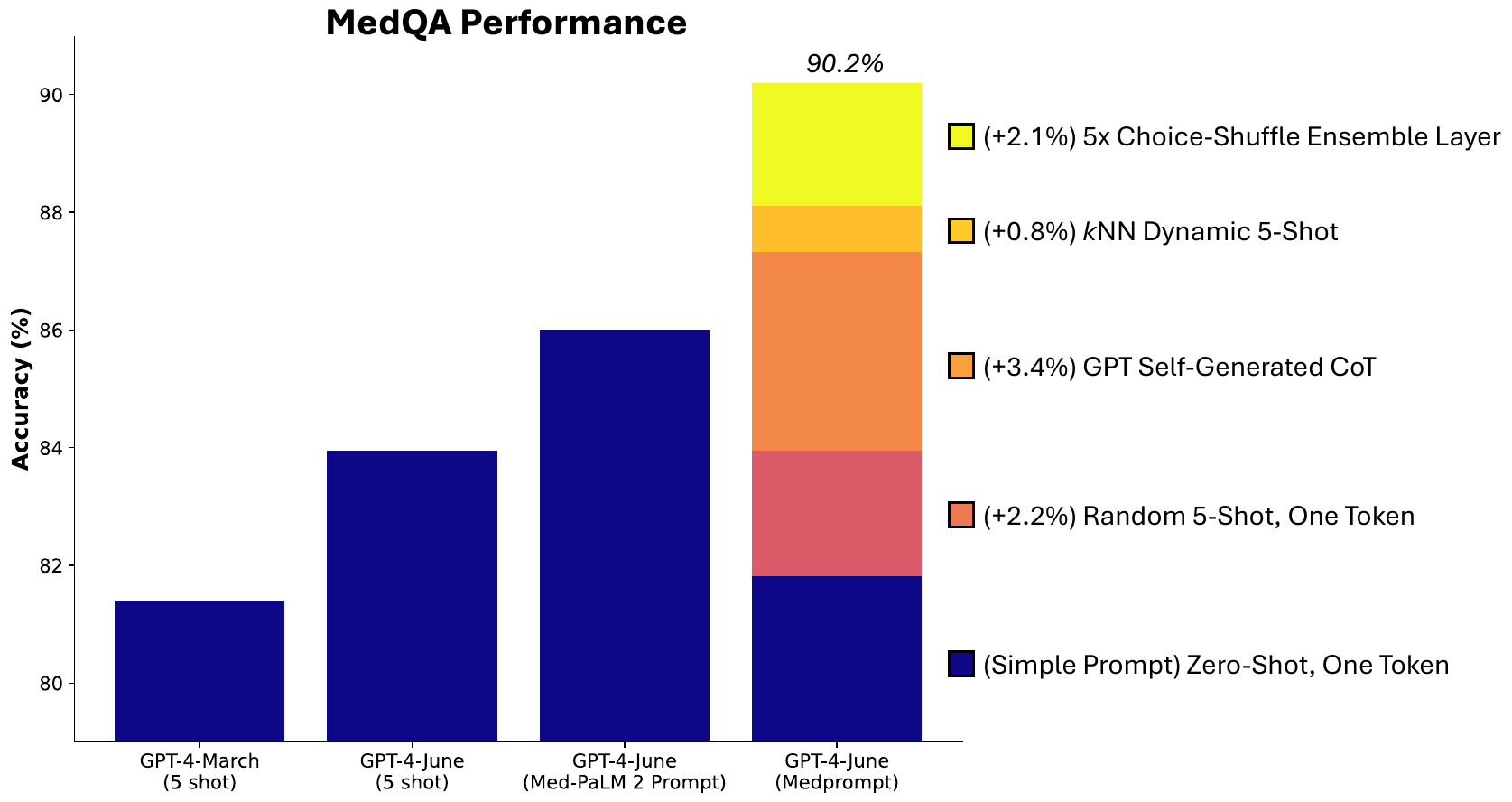}
    \caption{Identification of the relative contributions of different components of Medprompt via an ablation study.}
    \label{fig:ablation}
\end{figure}

Figure \ref{fig:ablation} shows the results of an ablation study conducted on the MedQA dataset, in an attempt to understand the relative contributions of each technique in Medprompt.
The blue bars represent prior work from \cite{nori2023capabilities}, and establish baselines for the Medprompt methodology. We then iteratively layered in each technique, and measured the relative difference in performance from each incremental change. 
As outlined in Section \ref{sec:combined-medprompt}, our base Medprompt strategy uses 5 kNN-curated few-shot exemplars and ensembles 5 API-calls together. 
We also experimented with setting up to 20 few-shot exemplars and up to 11 steps in the ensemble. 
We found that performance does increase marginally to 90.6\%, with additional few-shot exemplars and more ensemble steps. 
This suggests that further improvements on benchmarks may yet be possible, with a corresponding increase in inference time cost and complexity.
The introduction of chain-of-thought steps, as described in Section \ref{sec:prompting}, contributed the most to performance ($+3.4\%$), followed by few-shot prompting and choice shuffle ensembling ($+2.2\%$ each). 

The techniques we use are not statistically independent -- therefore, the order in which we test the contribution of each method matters.
Our choice of ordering for this ablation study is subjective and based on the relative complexity of the technique introduced.
A more theoretically sound method for credit allocation in the ablation study would involve the calculation of game-theoretic Shapley values \cite{shapley1953value}, which takes exponentially more model evaluations to test every potential permutation of orderings. 
We leave this to future work and encourage readers to think of the specific numbers in the ablation studies as reasonable approximations of relative contributions. 

\begin{table}[H]
\begin{center}
\caption{Ablation study on expert-crafted chain-of-thought (CoT) vs. GPT-4 self-generated CoT. Both use fixed 5-shot examples, with no ensemble.}
\vspace{5pt}
    
    \begin{tabular}{ll}
    \toprule
         &  \multicolumn{1}{c}{\textbf{MedQA}} \\
         &  US (4-option) \\ \midrule
        Expert-crafted CoT prompt from \cite{singhal2023expertlevel} & 83.8  \\
        GPT-4's self-generated CoT prompt & 86.9 (+3.1) \\ \bottomrule 
    \end{tabular}
    \label{tab:ablation-cot}
\end{center}
\end{table}

Apart from the stack of incremental changes, we compare the expert-crafted chain-of-thought (CoT) prompt used in Med-PaLM 2~\cite{singhal2023expertlevel} with the CoT prompt automatically generated by GPT-4 (Section \ref{sec:cot}). We evaluate GPT-4 using both prompts, with fixed 5-shot examples, no ensemble. Table \ref{tab:ablation-cot} reports their accuracy on the MedQA dataset. GPT-4's self-generated CoT outperforms the expert-crafted one by 3.1 absolute points. We notice that compared with the expert-crafted CoT used in Med-PaLM 2, CoT rationales generated by GPT-4 are longer and provide finer-grained step-by-step reasoning logic. One potential explanation is that GPT-4 generated CoT may be better suited to the model's own strengths and limitations, which could lead to improved performance when compared to the expert-crafted one. Another potential explanation is that expert-crafted CoT may contain implicit biases or assumptions that may not hold for all questions in the MedQA dataset, whereas GPT-4 generated CoT may be more neutral and generalizable across different questions.

\subsection{Generalization: Cross-Domain Exploration of Medprompt}
\label{sec:generalization}

We argue that the composition of prompt engineering techniques employed in Medprompt, based on a combination of dynamic few shot selection, self-generated chain of thought, and choice shuffle ensembling, have general purpose application. They are not custom-tailored to the MultiMedQA benchmark datasets.  To validate this, we further tested the final Medprompt methodology on six additional, diverse datasets from the MMLU benchmark suite covering challenge problems in the following subjects: electrical engineering, machine learning, philosophy, professional accounting, professional law, and professional psychology.
We further sourced two additional datasets answering NCLEX (National Council Licensure Examination) style questions, the exam required to practice as a registered nurse in the United States.

\begin{figure}[H]
    \centering
    \includegraphics[width=0.6\textwidth]{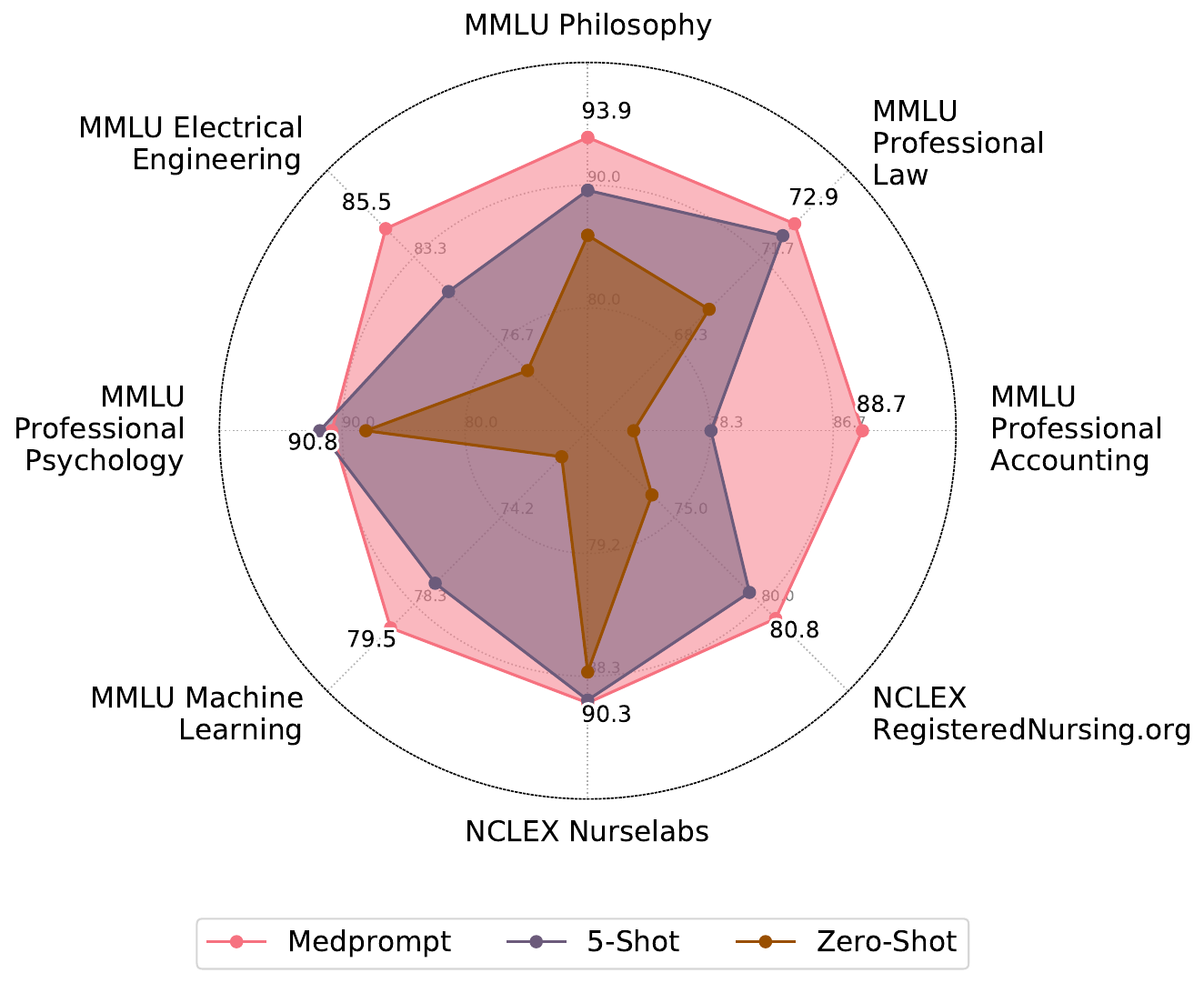}
    \caption{GPT-4 performance with three different prompting strategies on out of domain datasets. Zero-shot and five-shot approaches represent baselines and mirror the methodology followed in \cite{nori2023capabilities}.}
    \label{fig:ood-datasets}
\end{figure}

Figure \ref{fig:ood-datasets} shows GPT-4's performance on these diverse, out of domain dataset with Medprompt alongside zero-shot and five-shot prompts (with random exemplar selection).
Across these datasets, Medprompt provides an average improvement of $+7.3\%$ over baseline zero-shot prompting.
By comparison, Medprompt provided a $+7.1\%$ improvement over the same zero-shot baseline on the MultiMedQA datasets studied in this paper.
We emphasize that the similarity of improvement across datasets from different distributions demonstrates the generality of the Medprompt approach.
While beyond the scope of this paper, we believe the general framework underlying MedPrompt---a combination of few shot learning and chain-of-thought reasoning wrapped in an ensemble layer---can further generalize in applicability beyond the multiple choice question/answer setting with minor algorithmic modifications. For example, in an open-text generation setting, the ensemble layer may not be able to rely on a direct majority vote, but instead may aggregate by selecting the answer closest to all other answers in an embedding space. Another option would be to concatenate each of the $K$ generated pieces of text in a structured format and ask the model to select the most likely option, in the style of Ensemble Refinement \cite{singhal2023expertlevel}. We leave as future work exploration of the space of algorithmic modifications to other settings.

\section{Limitations and Risks}

Our paper highlights the power of systematic prompt engineering for steering generalist foundation models to amplify the specialist abilities of GPT-4 on medical challenge problems. We now share reflections on limitations and future directions from our assessment.  

As foundation models are trained on massive, internet-scale datasets, strong performance on benchmark problems may be due to memorization or leakage effects, where direct test samples have previously been observed by the model during training. In our previous study, which assessed the performance of GPT-4 on the datasets studied in this work with basic prompting~\cite{nori2023capabilities}, we introduced and ran a blackbox testing algorithm (MELD) which was unable to discover evidence of memorization. However, blackbox testing approaches like MELD are unable to guarantee that data has not been seen before. We also separately assessed GPT-4's performance on USMLE questions that were behind a paywall and, thus, not available on the public internet, and saw similarly strong performance \cite{nori2023capabilities}. In this study, we adopted standard machine learning best practices to control for overfitting and leakage during the prompt engineering process (Section \ref{sec:overfitting}). However, concerns of benchmark contamination during training remain.

Further, we note that the strong performance of GPT-4 with Medprompt cannot be taken to demonstrate real-world efficacy of the model and methods on open-world healthcare tasks \cite{nori2023capabilities}. While we are excited about the ability to steer foundations models to become top specialists on the benchmarks, we are cautious about taking the performance of the prompting strategies and model output to mean that the methods will be valuable in the practice of medicine in the open world, whether for automated or assisting healthcare professionals with administrative tasks, clinical decision support, or patient engagement in the open world. To be clear, the medical challenge problems that we and others have studied are designed for testing human competencies in selected domains. Such competency tests are typically framed as sets of multiple choice questions. Although such challenge problems are a common evaluation method and cover diverse topics, they do not capture the range and complexity of medical tasks that healthcare professionals face in actual practice. Thus, the pursuit of tests as proxies for real-world competency and the focus on multiple-choice style answers are limitations when it comes to transferring strong performance on speciality benchmarks to real-world performance. Futhermore, while we believe that the MedPrompt strategy can be adapted to non-multiple choice settings, we did not explicitly test these proposed adaptations on benchmarks in this work. 

We note that foundation models can generate erroneous information (sometimes referred to as \emph{hallucinations}) which may compromise generations and advice. While improvements in prompting strategies may lead to reductions in hallucinations and better overall accuracy, they may also make any remaining hallucinations even harder to detect. Promising directions include efforts on probabilistic calibration of generations, providing end-users with trustworthy measures of confidence in output.  In our prior study, we found that GPT-4 was well-calibrated and could provide trustable measures of its confidence on multiple choice test questions \cite{nori2023capabilities}.

We must also remain aware of biases in the output of foundation models. We do not yet understand how optimization in pursuit of top-level performance could influence other goals, such as equitable performance. It is vital to balance the pursuit of overall accuracy with equitable performance across different subpopulations to avoid exacerbating existing disparities in healthcare. Prior work has highlighted the need to understand and address biases in AI systems. The challenge of bias and fairness remains relevant and pressing in the context of model optimization, fine-tuning, and prompt engineering \cite{Howard2017, madaio2022, Zack2023}.  
\section{Summary and Conclusions}

We presented background, methods, and results of a study of the power of prompting to unleash top-performing specialist capabilities of GPT-4 on medical challenge problems, without resorting to special fine-tuning nor reliance on human specialist expertise for prompt construction. We shared best practices for evaluating performance, including the importance of evaluating model capabilities on an eyes-off dataset. We reviewed a constellation of prompting strategies and showed how they could be studied and combined via a systematic exploration. We found a significant amount of headroom in boosting specialist performance via steering GPT-4 with a highly capable and efficient prompting strategy. 

We described the composition of a set of prompting methods into Medprompt, the best performing prompting strategy we found for steering GPT-4 on medical challenge problems.  We showed how Medprompt can steer GPT-4 to handily top existing charts for all standard medical question-answering datasets, including the performance by Med-PaLM 2, a specialist model built via fine-tuning with specialist medical data and guided with handcrafted prompts authored by expert clinicians. Medprompt unlocks specialty skills on MedQA delivering significant gains in accuracy over the best performing model to date, surpassing 90\% for the first time on the benchmark. 

During our exploration, we found that GPT-4 can be tasked with authoring sets of custom-tailored chain-of-thought prompts that outperform hand-crafted expert prompts. We pursued insights about the individual contributions of the distinct components of the Medprompt strategy via ablation studies that demonstrate the relative importance of each component. We set aside eyes-off evaluation case libraries to avoid overfitting and found that the strong results by Medprompt are not due to overfitting. We explored the generality of Medprompt via performing studies of its performance on a set of competency evaluations in six fields outside of medicine, including electrical engineering, machine learning, philosophy, accounting, law, nursing, and clinical psychology. The findings in disparate fields suggests that Medprompt and its derivatives will be valuable in unleashing specialist capabilities of foundation models for numerous disciplines.  We see further possibilities for refining prompts to unleash speciality capabilities from generalist foundation models, particularly in the space of adapting the general MedPrompt strategy to non multiple choice questions. For example, we see an opportunity to build on the Medprompt strategy of using GPT-4 to compose its own powerful chain of thought examples and then employ them in prompting. Research directions moving forward include further investigation of the abilities of foundation models to reflect about and compose few-shot examples and to weave these into prompts.

While our investigation focuses on exploring the power of prompting generalist models, we believe that fine-tuning, and other methods of making parametric updates to foundation models are important research avenues to explore, and may offer synergistic benefits to prompt engineering. We maintain that both approaches should be judiciously explored for unleashing the potential of foundation models in high-stakes domains like healthcare.

\section*{Acknowledgments}

We thank S{\'e}bastien Bubeck, Peter Durlach, Peter Lee, Matthew Lungren, Satya Nadella, Joe Petro, Kevin Scott, Desney Tan, and Paul Vozila for discussion and feedback.

\bibliography{custom}

\begin{thebibliography}{10}

\bibitem{doi:10.1177/002224378402100210}
Niels~J. Blunch.
\newblock Position bias in multiple-choice questions.
\newblock {\em Journal of Marketing Research}, 21(2):216--220, 1984.

\bibitem{BioMedLM}
Elliot Bolton, David Hall, Michihiro Yasunaga, Tony Lee, Chris Manning, and
  Percy Liang.
\newblock Biomedlm, 2022.
\newblock Stanford Center for Research on Foundation Models.

\bibitem{brown2020language}
Tom~B. Brown, Benjamin Mann, Nick Ryder, Melanie Subbiah, Jared Kaplan,
  Prafulla Dhariwal, Arvind Neelakantan, Pranav Shyam, Girish Sastry, Amanda
  Askell, Sandhini Agarwal, Ariel Herbert-Voss, Gretchen Krueger, Tom Henighan,
  Rewon Child, Aditya Ramesh, Daniel~M. Ziegler, Jeffrey Wu, Clemens Winter,
  Christopher Hesse, Mark Chen, Eric Sigler, Mateusz Litwin, Scott Gray,
  Benjamin Chess, Jack Clark, Christopher Berner, Sam McCandlish, Alec Radford,
  Ilya Sutskever, and Dario Amodei.
\newblock Language models are few-shot learners, 2020.

\bibitem{bubeck2023sparks}
S{\'e}bastien Bubeck, Varun Chandrasekaran, Ronen Eldan, Johannes Gehrke, Eric
  Horvitz, Ece Kamar, Peter Lee, Yin~Tat Lee, Yuanzhi Li, Scott Lundberg,
  et~al.
\newblock Sparks of artificial general intelligence: Early experiments with
  gpt-4.
\newblock {\em arXiv preprint arXiv:2303.12712}, 2023.

\bibitem{caruana2004ensemble}
Rich Caruana, Alexandru Niculescu-Mizil, Geoff Crew, and Alex Ksikes.
\newblock Ensemble selection from libraries of models.
\newblock In {\em Proceedings of the twenty-first international conference on
  Machine learning}, page~18, 2004.

\bibitem{chowdhery2022palm}
Aakanksha Chowdhery, Sharan Narang, Jacob Devlin, Maarten Bosma, Gaurav Mishra,
  Adam Roberts, Paul Barham, Hyung~Won Chung, Charles Sutton, Sebastian
  Gehrmann, et~al.
\newblock Palm: Scaling language modeling with pathways.
\newblock {\em arXiv preprint arXiv:2204.02311}, 2022.

\bibitem{fernando2023promptbreeder}
Chrisantha Fernando, Dylan Banarse, Henryk Michalewski, Simon Osindero, and Tim
  Rocktäschel.
\newblock Promptbreeder: Self-referential self-improvement via prompt
  evolution.
\newblock {\em arXiv preprint arXiv:2309.16797}, 2023.

\bibitem{feurer2019hyperparameter}
Matthias Feurer and Frank Hutter.
\newblock Hyperparameter optimization.
\newblock {\em Automated machine learning: Methods, systems, challenges}, pages
  3--33, 2019.

\bibitem{gero2023selfverification}
Zelalem Gero, Chandan Singh, Hao Cheng, Tristan Naumann, Michel Galley,
  Jianfeng Gao, and Hoifung Poon.
\newblock Self-verification improves few-shot clinical information extraction.
\newblock In {\em ICML 3rd Workshop on Interpretable Machine Learning in
  Healthcare (IMLH)}, 2023.

\bibitem{gu2021pubmedbert}
Yu~Gu, Robert Tinn, Hao Cheng, Michael Lucas, Naoto Usuyama, Xiaodong Liu,
  Tristan Naumann, Jianfeng Gao, and Hoifung Poon.
\newblock Domain-specific language model pretraining for biomedical natural
  language processing.
\newblock {\em ACM Transactions on Computing for Healthcare (HEALTH)},
  3(1):1--23, 2021.

\bibitem{hendrycks2020measuring}
Dan Hendrycks, Collin Burns, Steven Basart, Andy Zou, Mantas Mazeika, Dawn
  Song, and Jacob Steinhardt.
\newblock Measuring massive multitask language understanding.
\newblock {\em arXiv preprint arXiv:2009.03300}, 2020.

\bibitem{hoffmann2022training}
Jordan Hoffmann, Sebastian Borgeaud, Arthur Mensch, Elena Buchatskaya, Trevor
  Cai, Eliza Rutherford, Diego de~Las Casas, Lisa~Anne Hendricks, Johannes
  Welbl, Aidan Clark, et~al.
\newblock Training compute-optimal large language models.
\newblock {\em arXiv preprint arXiv:2203.15556}, 2022.

\bibitem{Howard2017}
Ayanna~M. Howard, Cha Zhang, and Eric Horvitz.
\newblock Addressing bias in machine learning algorithms: {A} pilot study on
  emotion recognition for intelligent systems.
\newblock {\em 2017 IEEE Workshop on Advanced Robotics and its Social Impacts
  (ARSO)}, pages 1--7, 2017.

\bibitem{jin2021disease}
Di~Jin, Eileen Pan, Nassim Oufattole, Wei-Hung Weng, Hanyi Fang, and Peter
  Szolovits.
\newblock What disease does this patient have? a large-scale open domain
  question answering dataset from medical exams.
\newblock {\em Applied Sciences}, 11(14):6421, 2021.

\bibitem{jin2019pubmedqa}
Qiao Jin, Bhuwan Dhingra, Zhengping Liu, William~W Cohen, and Xinghua Lu.
\newblock Pubmedqa: A dataset for biomedical research question answering.
\newblock {\em arXiv preprint arXiv:1909.06146}, 2019.

\bibitem{ko-etal-2020-look}
Miyoung Ko, Jinhyuk Lee, Hyunjae Kim, Gangwoo Kim, and Jaewoo Kang.
\newblock Look at the first sentence: Position bias in question answering.
\newblock In {\em Proceedings of the 2020 Conference on Empirical Methods in
  Natural Language Processing (EMNLP)}, pages 1109--1121, Online, November
  2020. Association for Computational Linguistics.

\bibitem{kung2023performance}
Tiffany~H Kung, Morgan Cheatham, Arielle Medenilla, Czarina Sillos, Lorie
  De~Leon, Camille Elepa{\~n}o, Maria Madriaga, Rimel Aggabao, Giezel
  Diaz-Candido, James Maningo, et~al.
\newblock Performance of chatgpt on usmle: Potential for ai-assisted medical
  education using large language models.
\newblock {\em PLoS digital health}, 2(2):e0000198, 2023.

\bibitem{liu2021makes}
Jiachang Liu, Dinghan Shen, Yizhe Zhang, Bill Dolan, Lawrence Carin, and Weizhu
  Chen.
\newblock What makes good in-context examples for gpt-$3$?, 2021.

\bibitem{luo2022biogpt}
Renqian Luo, Liai Sun, Yingce Xia, Tao Qin, Sheng Zhang, Hoifung Poon, and
  Tie-Yan Liu.
\newblock Biogpt: generative pre-trained transformer for biomedical text
  generation and mining.
\newblock {\em Briefings in Bioinformatics}, 23(6):bbac409, 2022.

\bibitem{madaio2022}
Michael Madaio, Lisa Egede, Hariharan Subramonyam, Jennifer Wortman~Vaughan,
  and Hanna Wallach.
\newblock Assessing the fairness of ai systems: {AI} practitioners’
  processes, challenges, and needs for support.
\newblock In {\em 25th ACM Conference on Computer-Supported Cooperative Work
  and Social Computing (CSCW 2022)}, February 2022.

\bibitem{mccarthy2006proposal}
John McCarthy, Marvin~L Minsky, Nathaniel Rochester, and Claude~E Shannon.
\newblock A proposal for the {D}artmouth summer research project on artificial
  intelligence, {A}ugust 31, 1955.
\newblock {\em AI magazine}, 27(4):12--12, 2006.

\bibitem{newellshawsimonGPS1959}
Allen Newell, John~C Shaw, and Herbert~A Simon.
\newblock Report on a general problem solving program.
\newblock In {\em IFIP congress}, volume 256, page~64. Pittsburgh, PA, 1959.

\bibitem{nori2023capabilities}
Harsha Nori, Nicholas King, Scott~Mayer McKinney, Dean Carignan, and Eric
  Horvitz.
\newblock Capabilities of {GPT-4} on medical challenge problems.
\newblock {\em arXiv preprint arXiv:2303.13375}, 2023.

\bibitem{OpenAI2023GPT4TR}
OpenAI.
\newblock Gpt-4 technical report.
\newblock {\em ArXiv}, abs/2303.08774, 2023.

\bibitem{pal2022medmcqa}
Ankit Pal, Logesh~Kumar Umapathi, and Malaikannan Sankarasubbu.
\newblock Medmcqa: A large-scale multi-subject multi-choice dataset for medical
  domain question answering.
\newblock In {\em Conference on Health, Inference, and Learning}, pages
  248--260. PMLR, 2022.

\bibitem{pezeshkpour2023large}
Pouya Pezeshkpour and Estevam Hruschka.
\newblock Large language models sensitivity to the order of options in
  multiple-choice questions.
\newblock {\em arXiv preprint arXiv:2308.11483}, 2023.

\bibitem{schaeffer2023emergent}
Rylan Schaeffer, Brando Miranda, and Sanmi Koyejo.
\newblock Are emergent abilities of large language models a mirage?, 2023.

\bibitem{shapley1953value}
Lloyd~S Shapley et~al.
\newblock A value for n-person games.
\newblock 1953.

\bibitem{singhal2022large}
Karan Singhal, Shekoofeh Azizi, Tao Tu, S~Sara Mahdavi, Jason Wei, Hyung~Won
  Chung, Nathan Scales, Ajay Tanwani, Heather Cole-Lewis, Stephen Pfohl, et~al.
\newblock Large language models encode clinical knowledge.
\newblock {\em arXiv preprint arXiv:2212.13138}, 2022.

\bibitem{singhal2023expertlevel}
Karan Singhal, Tao Tu, Juraj Gottweis, Rory Sayres, Ellery Wulczyn, Le~Hou,
  Kevin Clark, Stephen Pfohl, Heather Cole-Lewis, Darlene Neal, Mike
  Schaekermann, Amy Wang, Mohamed Amin, Sami Lachgar, Philip Mansfield, Sushant
  Prakash, Bradley Green, Ewa Dominowska, Blaise~Aguera y~Arcas, Nenad Tomasev,
  Yun Liu, Renee Wong, Christopher Semturs, S.~Sara Mahdavi, Joelle Barral,
  Dale Webster, Greg~S. Corrado, Yossi Matias, Shekoofeh Azizi, Alan
  Karthikesalingam, and Vivek Natarajan.
\newblock Towards expert-level medical question answering with large language
  models, 2023.

\bibitem{sutskever2014seq2seqmt}
Ilya Sutskever, Oriol Vinyals, and Quoc~V Le.
\newblock Sequence to sequence learning with neural networks.
\newblock {\em Advances in neural information processing systems}, 27, 2014.

\bibitem{wang2023selfconsistency}
Xuezhi Wang, Jason Wei, Dale Schuurmans, Quoc Le, Ed~Chi, Sharan Narang,
  Aakanksha Chowdhery, and Denny Zhou.
\newblock Self-consistency improves chain of thought reasoning in language
  models, 2023.

\bibitem{wei2022emergent}
Jason Wei, Yi~Tay, Rishi Bommasani, Colin Raffel, Barret Zoph, Sebastian
  Borgeaud, Dani Yogatama, Maarten Bosma, Denny Zhou, Donald Metzler, Ed~H.
  Chi, Tatsunori Hashimoto, Oriol Vinyals, Percy Liang, Jeff Dean, and William
  Fedus.
\newblock Emergent abilities of large language models.
\newblock {\em Transactions on Machine Learning Research}, 2022.
\newblock Survey Certification.

\bibitem{wei2023chainofthought}
Jason Wei, Xuezhi Wang, Dale Schuurmans, Maarten Bosma, Brian Ichter, Fei Xia,
  Ed~Chi, Quoc Le, and Denny Zhou.
\newblock Chain-of-thought prompting elicits reasoning in large language
  models, 2023.

\bibitem{yang2023large}
Chengrun Yang, Xuezhi Wang, Yifeng Lu, Hanxiao Liu, Quoc~V Le, Denny Zhou, and
  Xinyun Chen.
\newblock Large language models as optimizers.
\newblock {\em arXiv preprint arXiv:2309.03409}, 2023.

\bibitem{yasunaga2022dragon}
Michihiro Yasunaga, Antoine Bosselut, Hongyu Ren, Xikun Zhang, Christopher~D.
  Manning, Percy Liang, and Jure Leskovec.
\newblock Deep bidirectional language-knowledge graph pretraining.
\newblock In {\em Neural Information Processing Systems (NeurIPS)}, 2022.

\bibitem{yasunaga2022linkbert}
Michihiro Yasunaga, Jure Leskovec, and Percy Liang.
\newblock Linkbert: Pretraining language models with document links.
\newblock In {\em Association for Computational Linguistics (ACL)}, 2022.

\bibitem{Zack2023}
Travis Zack, Eric Lehman, Mirac Suzgun, Jorge~A. Rodriguez, Leo~Anthony Celi,
  Judy Gichoya, Dan Jurafsky, Peter Szolovits, David~W. Bates, Raja-Elie~E.
  Abdulnour, Atul~J. Butte, and Emily Alsentzer.
\newblock Coding inequity: Assessing gpt-4{\textquoteright}s potential for
  perpetuating racial and gender biases in healthcare.
\newblock {\em medRxiv}, 2023.

\bibitem{zheng2023large}
Chujie Zheng, Hao Zhou, Fandong Meng, Jie Zhou, and Minlie Huang.
\newblock Large language models are not robust multiple choice selectors, 2023.

\bibitem{zheng2023judging}
Lianmin Zheng, Wei-Lin Chiang, Ying Sheng, Siyuan Zhuang, Zhanghao Wu, Yonghao
  Zhuang, Zi~Lin, Zhuohan Li, Dacheng Li, Eric.~P Xing, Hao Zhang, Joseph~E.
  Gonzalez, and Ion Stoica.
\newblock Judging llm-as-a-judge with mt-bench and chatbot arena, 2023.

\end{thebibliography}
\bibliographystyle{plain}

\end{document}